%% file: main_arxiv.tex
\documentclass[11pt]{article}

\usepackage[final]{acl}

\usepackage{times}
\usepackage{latexsym}
\usepackage{microtype}
\usepackage{graphicx}
\usepackage{booktabs}
\usepackage{xcolor}   
\usepackage{multirow}
\usepackage{graphicx}
\usepackage{amsmath} 
\usepackage{makecell}
\usepackage[normalem]{ulem}
\usepackage{algorithm}
\usepackage{algpseudocode}
\usepackage{hyperref}
\usepackage{enumitem} 
\usepackage{comment}
\usepackage{booktabs}

\usepackage{array}

\usepackage{colortbl} 
\usepackage[most]{tcolorbox}
\newcolumntype{Y}{>{\raggedright\arraybackslash}X}
\definecolor{BetterGreen}{HTML}{1a7f37}
\definecolor{BetterBlue}{HTML}{1f75cb}
\definecolor{BetterYellow}{HTML}{f5a800}
\definecolor{BetterRed}{HTML}{e00000}

\definecolor{promptgray}{RGB}{235,235,235}
\newcommand{\gptlink}{%
  \href{https://openai.com}{%
    \includegraphics[height=8pt]{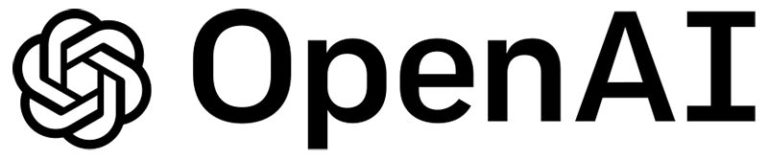}%
  }%
}
\newcommand{\geminilink}{%
  \href{https://deepmind.google/technologies/gemini/}{%
    \includegraphics[height=15pt]{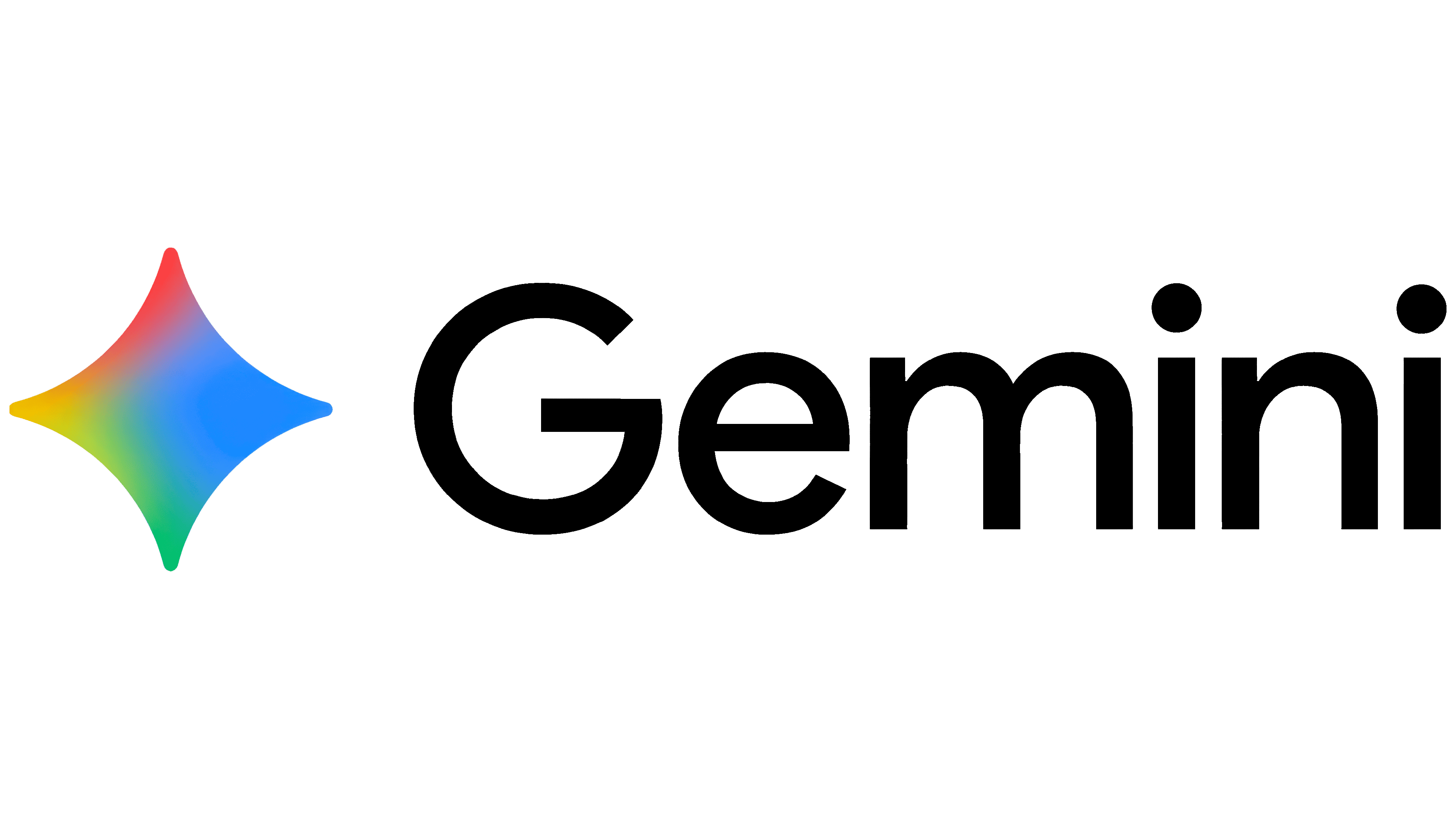}%
  }%
}

\usepackage[table]{xcolor}
\definecolor{bglightgray}{gray}{0.9}
\usepackage[T1]{fontenc}

\usepackage[utf8]{inputenc}

\usepackage{microtype}

\usepackage{inconsolata}

\usepackage{graphicx}

\title{From Experience to Skill: Multi-Agent Generative Engine \\ Optimization via Reusable Strategy Learning}

\author{
  \textbf{Beining Wu\textsuperscript{1, *}},
  \textbf{Fuyou Mao\textsuperscript{2, *}},
  \textbf{Jiong Lin\textsuperscript{1, *}},
  \textbf{Cheng Yang\textsuperscript{1, *}},
  \textbf{Jiaxuan Lu\textsuperscript{3}},
\\
  \textbf{Yifu Guo\textsuperscript{4},\textsuperscript{5}},
  \textbf{Siyu Zhang\textsuperscript{4}},
  \textbf{Yifan Wu\textsuperscript{6}},
  \textbf{Ying Huang\textsuperscript{1}},
  \textbf{Fu Li\textsuperscript{1, \dag}}
\\
\\
  \textsuperscript{1}Hangzhou Dianzi University,
  \textsuperscript{2}Central South University,  
\\  
  \textsuperscript{3}Shanghai Artificial Intelligence Laboratory,
  \textsuperscript{4}Sun Yat-sen University,
  \textsuperscript{5}Ramus,
\\  
  \textsuperscript{6}Hong Kong University of Science and Technology (GuangZhou)
\\
  \small{
    \textsuperscript{*}Equal contribution. \textsuperscript{\dag}Corresponding authors.
  }
\\
  \small{
    \textbf{Correspondence:} \href{mailto:lifu@hdu.edu.cn}{lifu@hdu.edu.cn}
  }
}

\begin{document}
\maketitle
\begin{abstract}
Generative engines (GEs) are reshaping information access by replacing ranked links with citation-grounded answers, yet current Generative Engine Optimization (GEO) methods optimize each instance in isolation, unable to accumulate or transfer effective strategies across tasks and engines. We reframe GEO as a strategy learning problem and propose MAGEO, a multi-agent framework in which coordinated planning, editing, and fidelity-aware evaluation serve as the execution layer, while validated editing patterns are progressively distilled into reusable, engine-specific optimization skills. To enable controlled assessment, we introduce a Twin Branch Evaluation Protocol for causal attribution of content edits and DSV-CF, a dual-axis metric that unifies semantic visibility with attribution accuracy. We further release MSME-GEO-Bench, a multi-scenario, multi-engine benchmark grounded in real-world queries. Experiments on three mainstream engines show that MAGEO substantially outperforms heuristic baselines in both visibility and citation fidelity, with ablations confirming that engine-specific preference modeling and strategy reuse are central to these gains, suggesting a scalable learning-driven paradigm for trustworthy GEO. Code is available at \href{https://github.com/Wu-beining/MAGEO}{https://github.com/Wu-beining/MAGEO}.
\end{abstract}

\section{Introduction}
\begin{figure*}[t]
    \centering
    \includegraphics[width=1.0\textwidth]{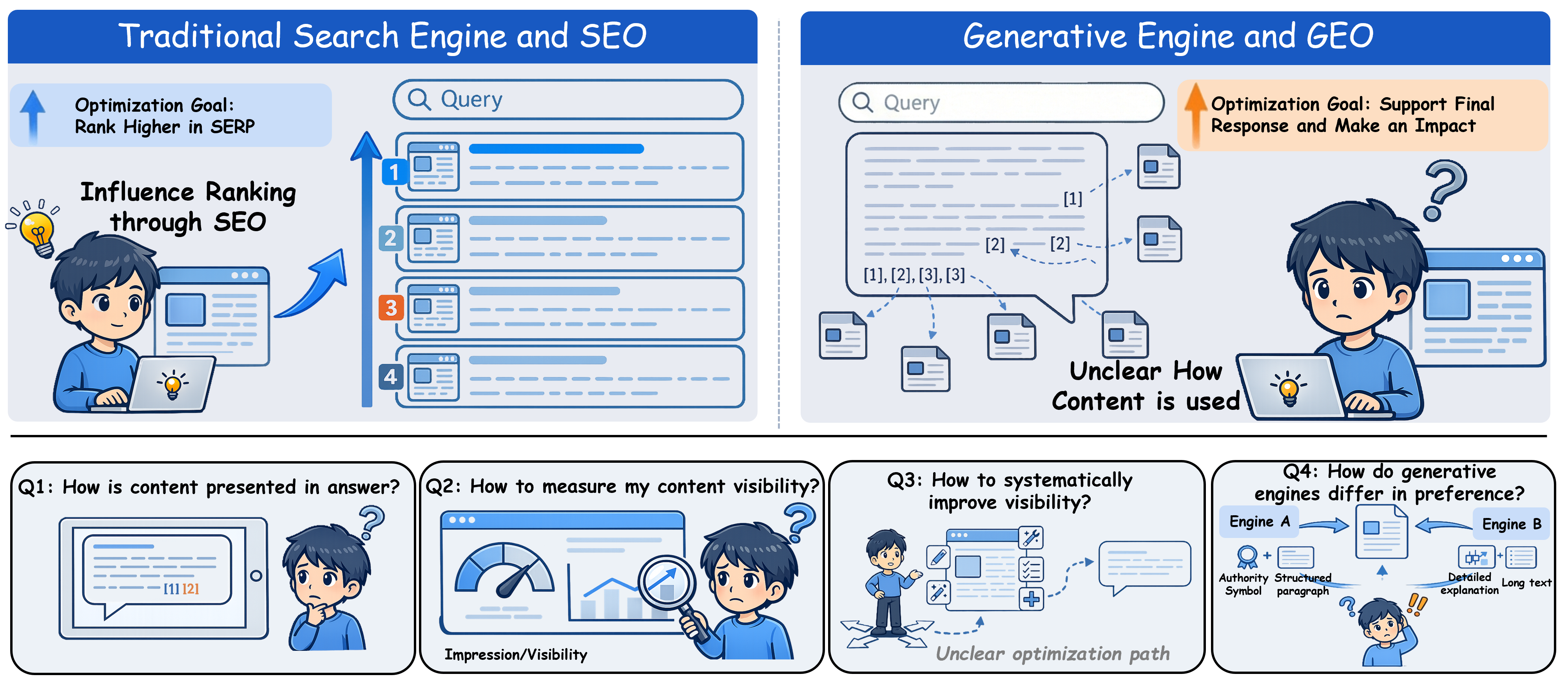}
    \caption{\textbf{The paradigm shift from SEO to GEO}. The transition from ranking-oriented goals to synthesis-based impact, highlighting four fundamental challenges: opaque presentation, undefined metrics, unclear optimization paths, and ambiguous preferences.}
    \label{fig1}
\end{figure*}
Recent advances in Large Language Models (LLMs) have accelerated the rise of Generative Engines (GEs) such as Gemini \cite{team2023gemini}, ChatGPT \cite{roumeliotis2023chatgpt}, and Qwen \cite{bai2023qwen}. Instead of returning ranked link lists, GEs typically use Retrieval-Augmented Generation (RAG) \cite{lewis2021retrievalaugmentedgenerationknowledgeintensivenlp} to retrieve evidence from multiple documents and generate answers with explicit citations~\cite{shi2024replug,asai2024selfrag}. This paradigm improves user efficiency but also reshapes creator-traffic dynamics: web pages increasingly serve as an evidential layer rather than the interaction endpoint, and ranking-based visibility alone no longer reflects actual impact.
 
For creators, this shift introduces opacity and a new optimization target. Retrieval, synthesis, and citation remain largely black-box processes, so creators cannot easily determine whether their content is used, ignored, or misattributed \cite{Godlevsky2017TheoreticalFO}. Traditional SEO signals \cite{sun2025ai}, such as keyword density and link structure, are often ineffective under semantically driven generation~\cite{yu2024rankrag,li2025searcho1}. Optimization must therefore move beyond search ranking toward improving citation accuracy and semantic influence within generated answers, a challenge central to Generative Engine Optimization (GEO). Crucially, effective GEO requires not only per-instance content improvement but also the ability to accumulate reusable optimization strategies that transfer across queries and engines.
 
Recent work has begun formalizing and evaluating GEO. GEO and GEO-Bench \cite{aggarwal2024geo} quantify exposure via position- and word-count-based measures alongside subjective impression ratings. RAID \cite{chen2025role} infers intent through staged planning and rewriting to align content with latent user needs. CC-GSEO-Bench \cite{chen2025ccgseobenchcontentcentricbenchmarkmeasuring} emphasizes the impact on answer quality, proposing dimensions including exposure, faithful credit, and causal impact.
 
However, several deployment-oriented gaps remain, as shown in Figure~\ref{fig1}. First, many metrics treat surface visibility and semantic influence separately without jointly enforcing faithful attribution, allowing exposure gains to coincide with miscitation or hallucination. Second, evaluations often rely on offline or semi-simulated pipelines where retrieval noise and ranking drift confound the effects of content edits~\cite{ru2024ragchecker,jin2025longrefiner}. Third, and most critically, all existing approaches optimize instances independently, with no mechanism to identify which editing patterns succeeded, abstract them into transferable strategies, or reuse them on subsequent tasks. Engine preference modeling also remains coarse~\cite{szymanski2025limitations}. As a result, current GEO remains trapped in per-instance trial-and-error rather than evolving into a cumulative, skill-building process~\cite{zhong2024memorybank,gupta2024metareflection,guo2025se}.
 
In this work, we reframe GEO as a strategy learning problem and propose MAGEO, a multi-agent framework that operates on two layers. At the \textit{execution layer}, four coordinated agents (Preference, Planner, Editor, and Evaluator) collaborate through an iterative Generate-Evaluate-Select loop in which the Evaluator enforces a fidelity gate and predicts DSV-CF gains to select the best candidate~\cite{liang2024mad,bo2024copper}. At the \textit{learning layer}, validated editing patterns are distilled into reusable, engine-specific optimization skills: within a session, effective trajectories and failures are recorded to guide subsequent rounds; across sessions, recurring successful patterns are abstracted into structured strategy skills indexed by engine and scenario for direct reuse~\cite{wang2025r3mem,zhong2024memorybank,yao2026toolace}. To enable controlled assessment, we introduce a Twin Branch Evaluation Protocol that compares generation with and without optimized content under identical retrieval lists, enabling causal attribution of edits in black-box engines. We further propose DSV-CF, a dual-axis metric that unifies semantic visibility with attribution accuracy while penalizing spurious citation, and construct MSME-GEO-Bench, a multi-scenario, multi-engine benchmark grounded in real-world queries across diverse life domains.
 
The main contributions are as follows:
 
\textbf{A multi-agent framework with reusable strategy learning.} We propose a dual-layer architecture in which multi-agent collaboration serves as the execution layer and strategy skill distillation serves as the learning layer. Validated editing patterns are abstracted into engine-specific skills that transfer across tasks, and ablations confirm that both engine-specific preference modeling and strategy reuse contribute measurably to the overall gains.
 
\textbf{Twin Branch protocol and DSV-CF for causal, fidelity-aware evaluation.} We introduce an instance-level controlled protocol that isolates the effect of content edits from retrieval variation, together with a dual-axis metric suite that jointly measures visibility and attribution quality while penalizing miscitation.
 
\textbf{MSME-GEO-Bench for multi-scenario, multi-engine research.} We release large-scale (Query, Engine, Source, Response) quadruples with scenario, intent, and complexity labels across diverse life domains and multiple mainstream engines, supporting robust cross-engine strategy evaluation.

\label{sec:method}
\begin{figure*}[t]
    \centering
    \includegraphics[width=1.0\textwidth]{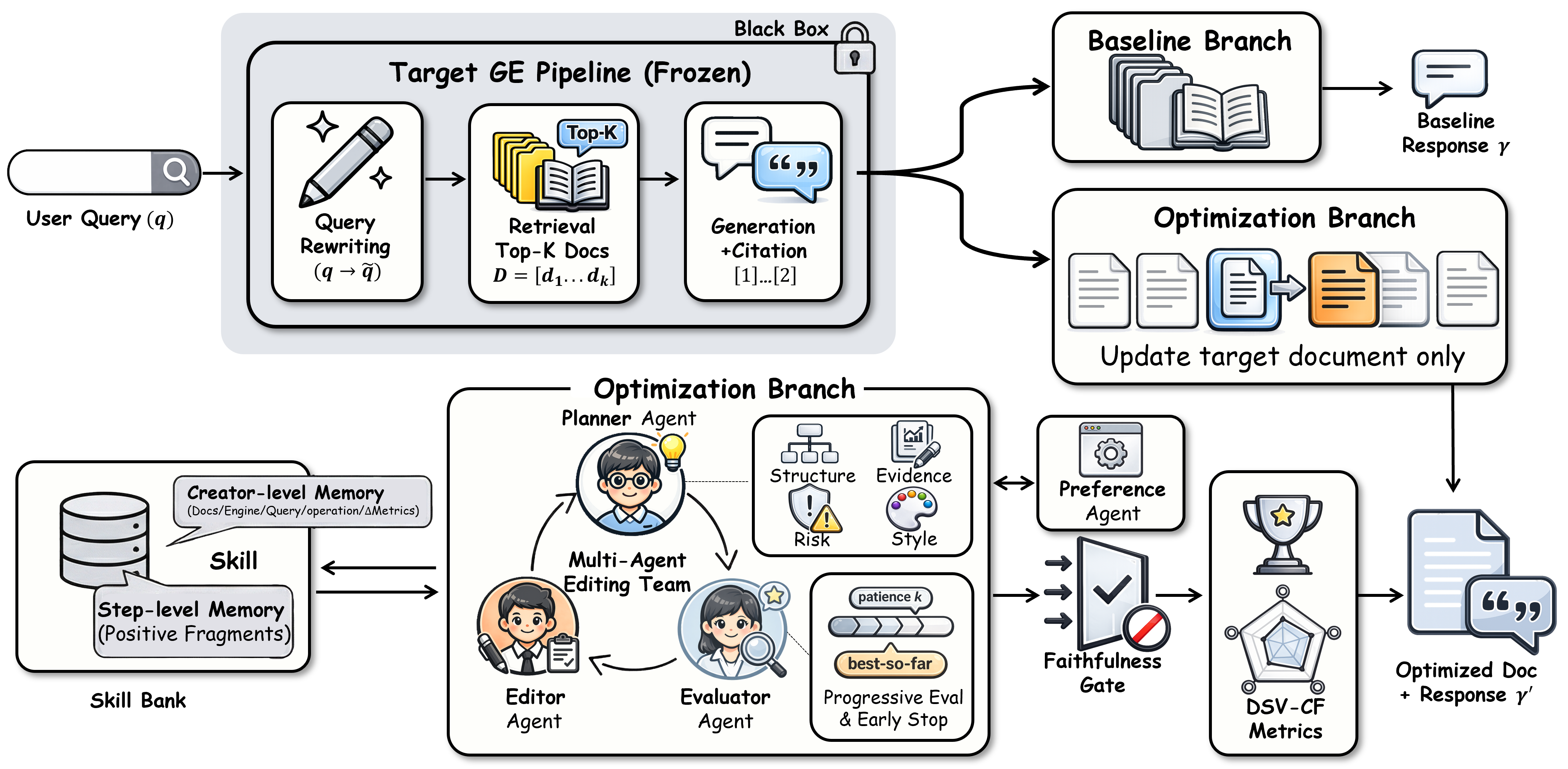}
    \caption{\textbf{Overview of MAGEO under the Twin-Branch protocol.} The upper panel compares the baseline branch and the optimization branch under the same frozen retrieval list. The lower panel is a detailed view of the optimization branch, showing how the Preference, Planner, Editor, and Evaluator agents interact with the Skill Bank.}
    \label{fig2}
\end{figure*}

\section{Related Work}

\subsection{Search Engine Optimization}
Classical information retrieval models retrieval as ranking candidate documents by relevance. This paradigm yields a mature toolkit in which users receive a ranked list and interact mainly by clicking links \cite{lindemann2025chatbots}.

Built around ranking on the search engine results page (SERP), SEO has been systematized as practices for improving page ranking and click-through rate. It typically distinguishes On-Page SEO, centered on content quality, structure and readability, from Off-Page SEO, which relies on link structure and site authority \cite{aggarwal2024geo}. Even when large models generate product descriptions and metadata at scale, optimization still targets observable signals such as keyword usage and link authority, with document-level ranking as the primary objective.

When search systems shift from returning links to producing natural language answers with citations, core assumptions of traditional SEO no longer hold. Evidence selection is implemented by a query rewriting–retrieval–generation pipeline rather than a transparent ranker, and visibility is reflected not only in page ranking but also in citation frequency, position and semantic role within answers. Existing studies indicate that keyword tuning and minor layout adjustments transfer poorly to semantically driven generative engines, motivating optimization frameworks explicitly tailored to this new paradigm \cite{chong-etal-2023-leveraging}.

\subsection{Retrieval-Augmented Language Models and Generative Engines}
With the rise of LLMs, RAG \cite{lewis2021retrievalaugmentedgenerationknowledgeintensivenlp} retrieves documents from external knowledge bases and feeds them as additional context so models can generate answers grounded in this evidence; it has become standard in open-domain and other knowledge-intensive question answering.

GEs further integrate retrieval and generation: instead of returning link lists, they aggregate retrieved evidence into cited, structured responses. Works such as GEO and AutoGEO abstract this behavior as a query rewriting--retrieval--generation pipeline and show that system outputs depend not only on retrieval quality but also on context selection and engine-specific preferences \cite{aggarwal2024geo,huang2025autogeo}.

Related research on conversational and agentic search models search as multi-turn dialogue or tool-using agents that iteratively plan, retrieve and reflect \cite{li2025deepthinkaligninglanguagemodels}. These studies illuminate how systems exploit retrieval and tools but usually treat web pages as interchangeable evidence rather than asking how a particular source document can strengthen its presence in generated results, thereby motivating creator-centered GEO.

GEO \cite{aggarwal2024geo} formulates creator-side optimization in black-box generative systems: internal engine parameters are fixed, and creators improve a page’s exposure and influence in generative answers only by editing the page itself. GEO-Bench pairs user queries with retrieved documents and introduces visibility metrics tailored to generative engines. Experiments show that simple strategies such as inserting explicit citations, adding key statistics, and emphasizing critical paragraphs substantially increase document visibility, whereas keyword stuffing in the style of traditional SEO is often ineffective or even harmful.

Subsequent work extends this framework along two main directions. RAID G-SEO explicitly models search intent in RAG-style black-box systems and uses staged summarization, intent inference and planned rewriting to better align pages with latent user needs. AutoGEO learns preference rules from generative engine behavior, distills them into natural language guidelines and applies them both via prompting and reward design. Across GEO-Bench and additional real-query benchmarks, these methods consistently improve visibility while largely preserving answer usefulness, indicating that intent-aware and preference-driven optimization is a promising basis for robust and cross-domain GEO.

\section{Methodology}

We reconfigure GEO from a heuristic modification paradigm into a controlled instance-level optimization process. To address the opacity of black-box engines, our framework adopts a strategy of freezing the retrieval context to decouple complex system interactions.

\subsection{Twin-Branch Evaluation Protocol}
\label{sec:protocol}
To scientifically isolate the causal impact of content optimization from retrieval ranking fluctuations, we formalize the problem as a twin-branch controlled experiment. Given a user query $q$ and a fixed retrieval list $\mathcal{L}_{ret}=\{d_{1},...,d_{K}\}$ obtained from a search engine, we define two parallel branches:
 
\noindent\textbf{Branch 1 (Baseline).} We maintain $\mathcal{L}_{ret}$ in its original state and employ the generative engine to produce a baseline response $r_{base}$.
 
\noindent\textbf{Branch 2 (Optimization).} We uniformly sample a target document $d_{target}$ from $\mathcal{L}_{ret}$ and apply semantic interventions to generate an optimized variant $d^*$. The retrieval list is updated in situ as $\mathcal{L}_{new}=\mathcal{L}_{ret}[d_{target}\leftarrow d^{*}]$. The engine then generates a response $r_{opt}$ based on this modified list.
 
The objective of MAGEO is to identify the optimal content variant $d^*$ that maximizes the comprehensive influence score $S$ in the generated response while preserving semantic fidelity:
\begin{equation}
d^{*}= \underset{d \in \Omega(d_{target})}{\text{argmax}} \;\; \mathcal{S}_{DSV-CF}(q, \mathcal{L}_{ret}[d_{target} \leftarrow d]),
\label{eq:ex1}
\end{equation}
where $\Omega(d_{target})$ denotes the space of candidate edits derived from the target document. This controlled protocol also provides the causal feedback signal for strategy learning: only when the effect of each edit is reliably attributed can the system determine which strategies are worth retaining as reusable skills.

\begin{figure*}[t]
    \centering
    \includegraphics[width=1.0\textwidth]{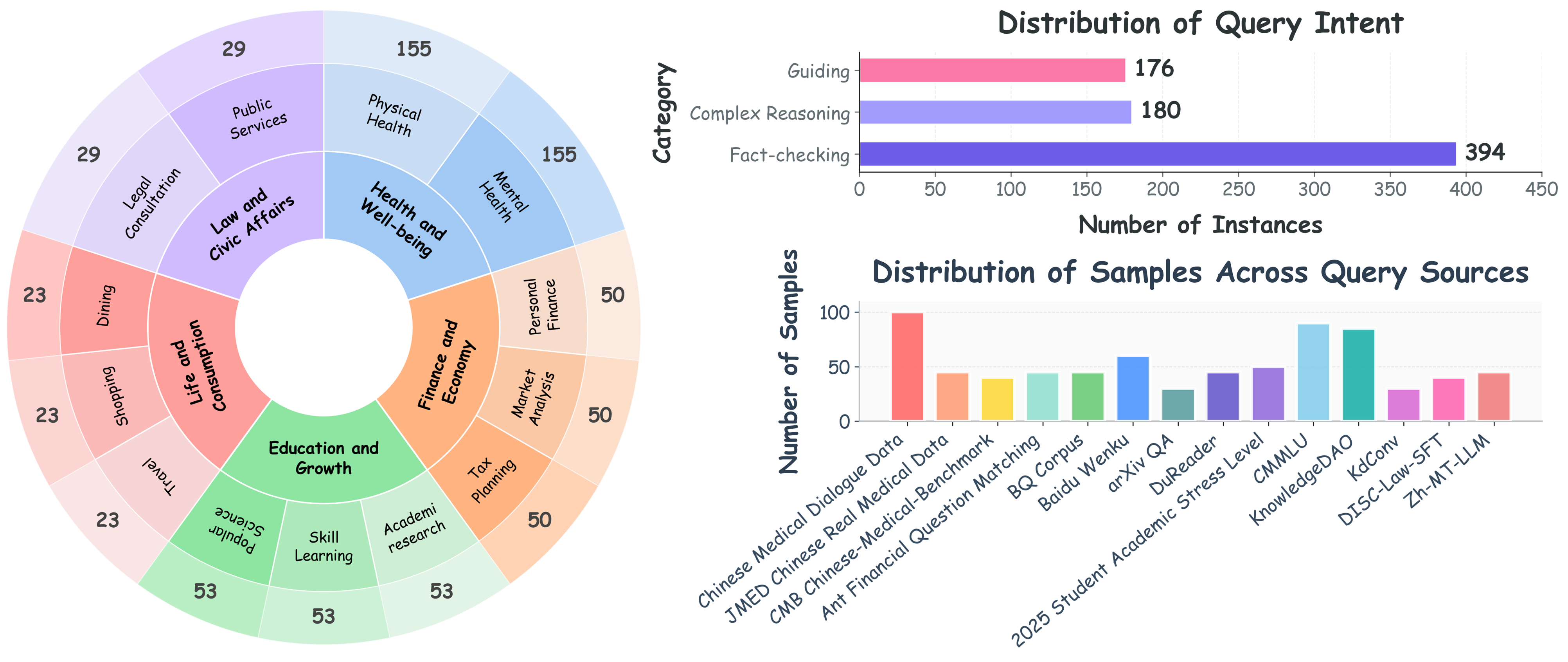}
    \caption{\textbf{Statistics analysis of MSME-GEO-Bench.} (left) Distribution of query scenarios. Our benchmark covers 5 major domains and 15 sub-category query types. (right) Distributions of query intent and sample sources. MSME-GEO-Bench incorporates a diverse array of user intents and data sources, enabling a comprehensive and multi-faceted evaluation of Generative Engine Optimization.}
    \label{fig:data_vis}
\end{figure*}

\subsection{The MAGEO Framework}
\label{sec:framework}
To solve the optimization problem defined in Eq.~\ref{eq:ex1}, MAGEO operates on two layers, as shown in Figure~\ref{fig2}. At the \textit{execution layer}, four specialized agents collaborate through a rigorous Generate-Evaluate-Select loop to iteratively optimize content \cite{yang2025multi,yang2026lungnoduleagent,yu2025visual}. At the \textit{learning layer}, validated editing patterns are consolidated into a Skill Bank for reuse on subsequent tasks.
 
\subsubsection{Multi-Agent Architecture}
\textbf{Preference Agent ($A_{pref}$).} This agent analyzes large-scale query-response quadruples to construct a Preference Profile $P_G$ for specific engines. It identifies tendencies such as statistical density, formatting preferences, and rhetorical patterns that are implicitly rewarded by different engines.
 
\noindent\textbf{Planner Agent ($A_{plan}$).} Acting as the editor-in-chief, the Planner synthesizes the engine profile $P_G$, the current response state, and relevant strategy skills retrieved from the Skill Bank to formulate high-level revision strategies. It decides \emph{what} should be improved, but does not directly edit the document.
 
\noindent\textbf{Editor Agent ($A_{edit}$).} The Editor executes the concrete modifications specified by $A_{plan}$. It generates candidate variants through parallel sampling, including structural adjustment, evidence enhancement, and style adaptation.
 
\noindent\textbf{Evaluator Agent ($A_{eval}$).} To reduce the latency of repeatedly calling the external engine, this agent functions as an internal quality inspector. It predicts DSV-CF gains using an LLM-as-a-Judge protocol and applies a \emph{Fidelity Gate}, rejecting any variant whose document-level semantic faithfulness falls below a threshold $\kappa$.
 
\subsubsection{Skill Bank}
The Skill Bank serves as the learning layer of MAGEO, transforming optimization experience into reusable strategy skills through a three-stage lifecycle: discovery, consolidation, and retrieval \cite{yu2026dual,bian2026realmem}.
 
\noindent\textbf{Step-level Memory ($M_S$).} This memory supports skill discovery within a single optimization session by recording the outcome of each editing attempt. Variants that produce positive DSV-CF gains are retained as positive fragments, while strategies that trigger fidelity or safety failures are flagged. These records constitute the raw material from which candidate skills are identified.
 
\noindent\textbf{Creator-level Memory ($M_C$).} This memory handles skill consolidation across optimization sessions. After a successful session, recurring effective patterns in $M_S$ are abstracted into structured strategy skills. Each skill is characterized by its applicability conditions (engine type, scenario), the editing operations it prescribes, and its observed effectiveness ($\Delta$Metrics). Skills are indexed by engine and scenario and stored in the Skill Bank for cross-instance reuse. To keep the Skill Bank scalable, we enforce a capacity limit for each engine-scenario combination and evict entries by recency or usage frequency once the limit is exceeded, motivated by decoupling and anti-forgetting strategies in continual learning ~\cite{guo2025decoupling,lu2026zero}.
 
\noindent\textbf{Skill Retrieval.} When a new optimization task arrives, $A_{plan}$ queries the Skill Bank with the current engine type and scenario to retrieve matching strategy skills. Retrieved skills serve as prior guidance for revision planning, narrowing the search space and reducing the exploration rounds required.
 
\noindent\textbf{Optimization Loop.} In each round $t$, $A_{plan}$ retrieves relevant skills from the Skill Bank and current constraints from $M_S$ to guide $A_{edit}$. The Editor produces a candidate pool $V_t$. The Evaluator filters $V_t$ using the fidelity gate and selects the best surviving variant $d_{t+1}$ based on predicted gains. The loop terminates when the score plateaus or the edit budget is exhausted. Upon termination, $M_S$ is consolidated into the Skill Bank via skill consolidation, completing the learning cycle.

\section{MSME-GEO-Bench}
\subsection{Dataset Construction}
\label{sec:dataset_construction}

We construct \textbf{MSME-GEO-Bench} to improve query--document alignment and coverage of everyday scenarios. The benchmark is grounded in ELIS theory~\citep{Savolainen2010EverydayLI} and organized by the HLD-QT taxonomy to better reflect decision-oriented information seeking rather than simple factoid retrieval.

Our construction pipeline contains four stages. First, we create seed queries spanning the HLD-QT space, retrieve candidate documents with the Tavily Search API, keep the Top-10 results, and randomly lock one source document $d_{\mathrm{src}}$. We then use Gemini-3 Pro to reverse-generate user queries that $d_{\mathrm{src}}$ can answer, enforcing strong semantic alignment between the query and the selected source. Second, we perform strict closed-loop retrieval validation by re-submitting each generated query to Tavily and retaining it only if $d_{\mathrm{src}}$ still appears in the Top-10. This step ensures that the query--document link is observable under fixed retrieval and makes subsequent optimization effects measurable in a realistic black-box setting. Third, for each validated sample, Gemini-3 Pro assigns three labels: core life domain, interaction intent, and query complexity. As summarized in Figure~\ref{fig:data_vis}, the resulting benchmark covers 5 major domains and 15 sub-categories together with diverse intent types and source distributions. Finally, to reduce model-specific construction bias, we apply structured prompting, lightweight rule-based filtering, and sampled human quality checks. Manual inspection on the test split shows over 95\% tag precision.

\subsection{The DSV-CF Metric}
\label{sec:metric}

Existing evaluation metrics often fail to distinguish effective exposure from spurious citation. To address this, we propose the Dual-Axis Semantic Visibility and Content Fidelity (DSV-CF) framework, which contains two primary components:

 \textbf{Surface Semantic Visibility (SSV).} This component quantifies the exposure intensity. It aggregates Word-Level Visibility (WLV), Decayed Positional Authority (DPA), Citation Prominence (CP), and Subjective Impression (SI).
\textbf{Intrinsic Semantic Impact (ISI).} This component utilizes LLMs to assess the depth of influence. It includes Attribution Accuracy (AA), Response-level Faithfulness ($FA_{resp}$), Key-Point Coverage (KC), and Answer Dominance (AD).

We synthesize these dimensions into a single optimization objective:
\begin{equation}
S_{DSV-CF} = \lambda \cdot \bar{S}_{SSV} + (1-\lambda) \cdot \bar{S}_{ISI} - \gamma (1 - AA),
\label{eq:dsvcf}
\end{equation}

where $\bar{S}_{SSV}$ and $\bar{S}_{ISI}$ are the normalized aggregates of the sub-metrics. The hyperparameter $\lambda$ balances visibility and quality, while $\gamma$ controls the penalty severity for citation errors. In our experiments, we set $\lambda = 0.5$ to impose a symmetric prior between exposure gain and fidelity preservation: larger values tend to over-reward visibility-oriented edits, whereas smaller values reduce the task to conservative rewriting with limited competitive gain. We use $\gamma = 0.5$ as the default attribution penalty because it provides the best overall DSV-CF among representative settings on the test set.
\section{Experiments}

\input{table/main_result}

In this section, we answer five questions: (RQ1) Can MAGEO improve content visibility while preserving attribution fidelity across different generative engines? (RQ2) How well does the LLM-based DSV-CF judge align with human assessment? (RQ3) What is the cost--effectiveness trade-off of multi-agent GEO? (RQ4) How much do the Skill Bank and engine-specific preference modeling contribute? (RQ5) Does the multi-agent evolutionary process provide gains beyond simple combinations of heuristic strategies?

\subsection{Experimental Setup}

\paragraph{Datasets.}
We evaluate on two benchmarks. \textbf{MSME-GEO-Bench (ours)} is a multi-scenario benchmark comprising real-world user queries across health, finance, education, consumption, and related daily-life domains. \textbf{GEO-Bench} \cite{aggarwal2024geo} is the standard benchmark from prior GEO work and is included for direct comparison with previously published baselines.

\paragraph{Target Engines.}
We evaluate three representative engines. \textbf{Proprietary models:} GPT-5.2 (OpenAI) and Gemini-3 Pro (Google). \textbf{Open-weights model:} Qwen-3 Max, which represents strong open deployments in private search solutions.

\paragraph{Baselines.}
We compare MAGEO against the nine heuristic GEO strategies released in the official GEO repository \cite{aggarwal2024geo}: \textit{Authoritative}, \textit{Citing Credible Sources}, \textit{Statistics Addition}, \textit{Quotation Addition}, \textit{Easy-to-Read}, \textit{Fluent}, \textit{Unique Words}, \textit{Technical Terms}, and \textit{Keyword Optimization}. These are the only publicly released GEO baselines with fully reproducible implementations at the time of our experiments.

\paragraph{Metrics.}
We use DSV-CF as the optimization and evaluation metric. We also report the constituent sub-metrics to expose the trade-off among visibility, semantic transfer, and attribution fidelity.

\subsection{Human Validation of the LLM-as-a-Judge}
\label{sec:human_validation}

Because DSV-CF partially relies on LLM judgments, we validate it against human experts. We stratify and randomly sample 100 quadruple from MSME-GEO-Bench across different scenarios, intents, and complexity levels, including both original responses and MAGEO-optimized responses. Three annotators with NLP backgrounds independently read the query, a source-document summary, and the response, then score \emph{visibility}, \emph{semantic influence}, and \emph{attribution faithfulness} on a 1-10 scale. We linearly combine these three human scores using the same weights as DSV-CF and average over annotators to obtain a human counterpart of the metric.

\begin{table}[t]
\centering
\small
\caption{Human validation of the LLM-based DSV-CF judge on 100 sampled triplets.}
\label{tab:human_validation}
\begin{tabular}{lccc}
\toprule
Metrics & $\rho$ & 95\% CI & p-value \\
\midrule
DSV-CF & 0.81 & [0.76, 0.85] & $< 1\mathrm{e}{-10}$ \\
WLV    & 0.79 & [0.73, 0.84] & $< 1\mathrm{e}{-10}$ \\
CF     & 0.74 & [0.67, 0.80] & $< 1\mathrm{e}{-9}$ \\
\bottomrule
\end{tabular}
\end{table}

As shown in Table~\ref{tab:human_validation}, the LLM-based judge exhibits strong agreement with human evaluation on the overall metric and key sub-dimensions, with Spearman correlations of 0.81 for DSV-CF, 0.79 for WLV, and 0.74 for CF. In an additional pairwise comparison on 50 sampled response pairs, the agreement rate between the LLM judge and human experts reaches 81.5\%, significantly above random choice. These results suggest that the LLM judge is a reliable proxy for scalable GEO evaluation, although it remains an approximation and should be supplemented with sampled human audits in high-risk settings.

\begin{table}[t]
\centering
\resizebox{\columnwidth}{!}{
\begin{tabular}{lcccccccc}
\toprule
& \multicolumn{8}{c}{\textbf{Qwen-3 Max}} \\
\cmidrule(lr){2-9}
& \multicolumn{4}{c}{\textbf{SSV}} & \multicolumn{4}{c}{\textbf{ISI}} \\
\cmidrule(lr){2-5} \cmidrule(lr){6-9}
\textbf{Method} & WLV & DPA & CP & SI & AA & FA & KC & AD \\
\midrule
\multicolumn{9}{l}{\textit{Performance without Generative Engine Optimization}} \\
None & 1.00 & 1.00 & 1.33 & 6.21 & 6.37 & 5.82 & 5.61 & 6.12 \\
\midrule
\multicolumn{9}{l}{\textit{High-Performing Generative Engine Optimization Methods}} \\
Fluent & 0.66 & 0.66 & 4.91 & 6.43 & 6.50 & 6.41 & 5.91 & 5.34 \\
Unique Words & 0.69 & 0.71 & 4.96 & 6.33 & 6.08 & 5.91 & 5.74 & 5.59 \\
Authoritative & 1.10 & 1.10 & 4.62 & 6.39 & 6.15 & 6.32 & 5.92 & 5.65 \\
More Quotes & 1.33 & 1.16 & 4.79 & 6.40 & 6.49 & 6.49 & 5.99 & 5.54 \\
Citing Credible & 0.92 & 0.94 & 4.89 & 6.30 & 6.16 & 6.27 & 6.01 & 5.91 \\
Simple Language & 0.97 & 1.05 & 4.78 & 6.42 & 6.90 & 6.67 & 5.81 & 5.40 \\
Technical Terms & 0.75 & 0.75 & 4.55 & 6.35 & 6.07 & 6.16 & 6.08 & 5.64 \\
Stats Optimization & 0.78 & 0.80 & 4.81 & 6.40 & 5.99 & 5.92 & 5.92 & 5.75 \\
SEO Optimize & 0.74 & 0.74 & 4.77 & 6.18 & 6.28 & 5.92 & 5.86 & 5.51 \\
\midrule
\multicolumn{9}{l}{\textit{Multi-Agent GEO with Strategy Learning (Ours)}} \\
\textbf{Main (Ours)} & \textbf{3.84} & \textbf{3.84} & \textbf{5.89} & \textbf{6.65} & \textbf{6.77} & \textbf{6.94} & \textbf{6.67} & \textbf{6.41} \\
w/o Engine Preference & \underline{1.77} & \underline{1.79} & \underline{5.64} & \underline{6.60} & \underline{6.74} & \underline{6.77} & \underline{6.35} & \underline{5.98} \\
w/o Skill Bank & 1.20 & 1.33 & 5.54 & 6.32 & 6.56 & 6.48 & 6.08 & 5.88 \\
\bottomrule
\end{tabular}
}
\caption{Performance comparison on using Qwen-3 Max model. The best and second-best results in each column are \textbf{bolded} and \underline{underlined}, respectively.}
\label{tab:qwen_results}
\end{table}

\subsection{Main Results}

\paragraph{MAGEO Establishes New SOTA.}

As shown in Table~\ref{tab:main_result}, MAGEO consistently outperforms all single-heuristic baselines across both benchmark datasets. On MSME-GEO-Bench, MAGEO achieves a WLV of \textbf{4.52} with GPT-5.2, more than tripling the strongest baseline (\textit{More Quotes}, 1.33); on Gemini-3 Pro, this rises to \textbf{5.30}, substantially above the best single heuristic. Improvements are also consistent across CP, SI, AA, FA, KC, and AD. On GEO-Bench, the same pattern persists. These gains indicate that MAGEO is not merely increasing superficial exposure: it improves both semantic transfer and citation faithfulness.

\paragraph{Fidelity-Aware Optimization.}
MAGEO's visibility gains do not come from indiscriminate citation amplification. Methods such as \textit{Keyword Optimization} can trigger hallucination penalties by forcing lexical patterns that disrupt semantic coherence. In contrast, MAGEO maintains high fidelity ($FA_{doc} > 7.05$) while increasing visibility, indicating that the Evaluator Agent and fidelity gate filter harmful edits.

\paragraph{Generalization to an open-weights engine.}
We observe the same trend on Qwen-3 Max. Under the same evaluation pipeline, MAGEO increases WLV/DPA from 1.00/1.00 without GEO to 3.84/3.84, while maintaining strong fidelity-related scores (CP 5.89, SI 6.65, AA 6.77, FA 6.94, KC 6.67, and AD 6.41). In contrast, heuristic baselines on Qwen-3 Max remain around WLV $\leq$ 1.33, indicating that MAGEO is not limited to proprietary engines and also transfers to strong open deployments.

\subsection{Ablation Study}
We examine the contribution of key components in MAGEO (Table \ref{tab:main_result}, bottom rows).
 
\textbf{Impact of Engine-Specific Preference Modeling:} Removing the engine preference module causes a sharp performance drop ($\sim$19\% on GPT 5.2). This confirms that knowing the judge is critical; generic high-quality writing is insufficient for GEO. Successful optimization also requires alignment with engine-specific preferences learned by the Preference Agent.
 
\textbf{Impact of the Skill Bank:} Removing the Skill Bank results in a $\sim$13\% drop. Without accumulated strategy skills, the Planner Agent cannot leverage successful optimization patterns from previous instances (e.g., Gemini-3 Pro prefers bullet points for medical advice), reverting to trial-and-error which is less query-efficient.
 
These trends are also qualitatively consistent with cross-engine preference differences: Gemini-3 Pro tends to favor compact and highly structured evidence presentation, GPT-5.2 more often adopts an authority-seeking style with heavier citation formatting, and Qwen-3 Max prefers didactic, safety-aware organization.

\begin{figure}[t]
    \centering
    \includegraphics[width=1.0\columnwidth]{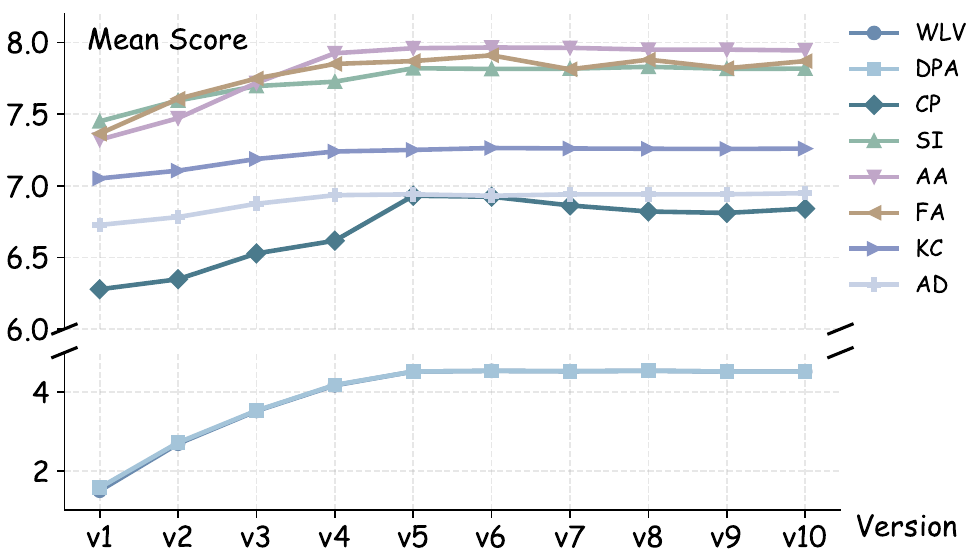}
    \caption{Evolutionary optimization trajectory of MAGEO, showing performance peaking at Version 5 before diminishing.} 
    \label{fig3}
\end{figure}

\begin{figure}[t]
    \centering
    \includegraphics[width=1.0\columnwidth]{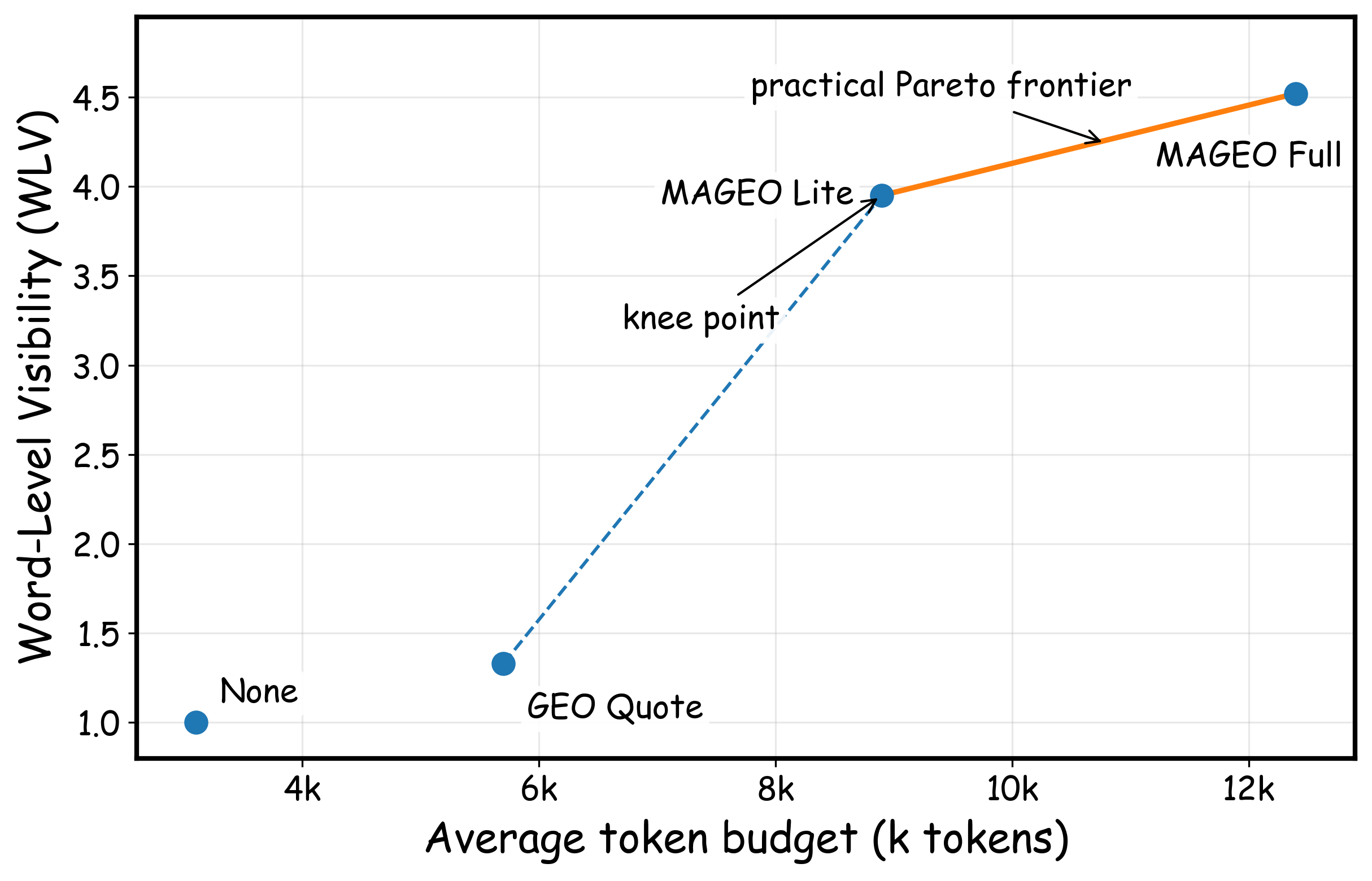}
    \caption{Cost-effectiveness trade-off of MAGEO on MSME-GEO-Bench with GPT-5.2. MAGEO Lite and MAGEO Full form the practical Pareto frontier, with Lite serving as the knee point and the more cost-effective default.}
    \label{fig:pareto_frontier}
\end{figure}

\label{sec:cost_analysis}

\subsection{Analysis of Evolutionary Optimization}
As shown in Figure~\ref{fig3}, the visibility score improves rapidly in the first few rounds and peaks around Version 5. Early gains mainly come from structural repair and missing-evidence completion, with the Skill Bank helping reduce early-round exploration cost by providing validated starting strategies. The peak corresponds to a good balance between information density and readability. Beyond this point, additional edits bring diminishing returns and can even slightly reduce faithfulness---a phenomenon we call \emph{over-optimization fatigue}. This observation motivates dynamic early stopping in MAGEO.

\begin{table}[t]
\centering
\small
\caption{Cost-effectiveness comparison on MSME-GEO-Bench with GPT-5.2.}
\label{tab:cost_analysis}
\begin{tabular}{lccc}
\toprule
Method & Avg. Tokens & WLV & Avg. Latency \\
\midrule
None & 3.1k (1.0$\times$) & 1.00 & -- \\
GEO Quote & 5.7k (1.8$\times$) & 1.33 & 12.3s \\
MAGEO Lite & 8.9k (2.9$\times$) & 3.95 & 18.1s \\
MAGEO Full & 12.4k (4.0$\times$) & 4.52 & 38.7s \\
\bottomrule
\end{tabular}
\end{table}

\subsection{Cost-Effectiveness Analysis}
A multi-agent GEO framework is only practically meaningful if its visibility gains are cost-effective relative to additional inference overhead. We therefore measure the total number of input/output tokens across all LLM calls for one query as a unified inference budget and compare four settings on MSME-GEO-Bench with GPT-5.2: \textit{None}, \textit{GEO Quote}, \textit{MAGEO Lite}, and \textit{MAGEO Full}.

Table~\ref{tab:cost_analysis} shows that both MAGEO variants dominate heuristic GEO in the visibility--cost space. \textbf{MAGEO Lite} uses ${\sim}2.9\times$ the tokens of GEO Quote but achieves nearly $3\times$ the WLV, while scaling to Full yields a smaller marginal gain (3.95$\rightarrow$4.52). We therefore recommend \textbf{Lite} for cost-sensitive and \textbf{Full} for peak-performance applications. Notably, the false-citation ratio (CF) falls from 0.058 (GEO Quote) to 0.047 (Lite) and 0.043 (Full), confirming gains are not driven by hallucinated citations. Paired t-tests confirm both variants significantly outperform GEO Quote ($p<1\mathrm{e}{-8}$), and Full outperforms Lite ($p<0.01$), though the additional gain is modest. Figure~\ref{fig:pareto_frontier} further illustrates that Lite already captures most of the achievable visibility gain \cite{wu2025concise}.

\input{table/comb_result}
 
\subsection{Comparison with Combinatorial Baselines}
\label{sec:combo_baselines}
A natural question is whether MAGEO is simply a more complex way to stack existing heuristics. To test this, we construct \textbf{Combo-Best}, which combines the strongest heuristic rules (More Quotes + Citing Source + Authoritative + Technical Terms) in a single pipeline.
 
As shown in Table~\ref{tab:comb_results}, Combo-Best is stronger than any individual heuristic but still remains substantially below MAGEO. This indicates that MAGEO's gains are not reducible to additive rule composition. The advantage comes from iterative, engine-aware, and fidelity-aware coordination across agents, combined with reusable strategy skills that guide optimization beyond what static rule composition can achieve.

\section{Conclusion}

We reframe GEO as a strategy learning problem and propose MAGEO, a multi-agent framework coupling iterative optimization with reusable skill distillation. Together with Twin Branch, DSV-CF, and MSME-GEO-Bench, it provides a unified pipeline from causal assessment to skill accumulation. Experiments on three mainstream engines confirm substantial gains over heuristic baselines in both visibility and citation fidelity, while ablations validate the contribution of engine-specific preference modeling and strategy reuse. These results suggest GEO is best approached not as ad hoc rule engineering but as a structured learning process in which optimization experience is consolidated into transferable skills. Future work will address multimodal GEO and adaptive skill maintenance that tracks engine distribution drift.

\section{Acknowledgements}
This paper is part of the 2025 Special Social Science Research Project of the Ministry of Education (Higher Education Counselors Research) entitled “Research on Mechanism Innovation of Systematic Transformation of Ideological and Political Education in Universities under the Background of
New AI Technology Application” (Project No. 25JDSZ3109)

\section*{Limitations}
Despite its strong empirical performance, MAGEO still has several limitations. First, its multi-agent optimization loop introduces higher token cost and latency than lightweight heuristic methods, which may limit deployment in real-time or high-throughput settings. Second, although MSME-GEO-Bench provides structured annotations and realistic generation logs, its current scale and category distribution still constrain fine-grained subgroup analysis. Third, because Gemini-3 Pro is involved in reverse query generation and annotation, the benchmark may retain some model-specific bias despite our retrieval validation, rule-based filtering, and sampled human checks. Fourth, while the Skill Bank demonstrates measurable gains in ablation, we do not yet provide a formal analysis of skill generalization across unseen scenarios or a learning curve showing how optimization efficiency scales with accumulated experience. Finally, as generative engines evolve over time, learned skills may gradually lose effectiveness, and our current framework is limited to text-only GEO rather than multimodal settings.

\bibliography{custom}

\appendix

\clearpage
\section{MSME-GEO-Bench Construction}
\label{sec:appendix_a}

To address the limitations of existing benchmarks-especially weak query-document alignment and limited coverage of everyday scenarios-we construct \textbf{MSME-GEO-Bench}, a \textbf{M}ulti-\textbf{S}cenario, \textbf{M}ulti-\textbf{E}ngine GEO benchmark grounded in Everyday Life Information Seeking (ELIS) theory.

\subsection{Theoretical Grounding: ELIS}
Unlike prior benchmarks that rely on coarse labels such as health or finance alone, MSME-GEO-Bench is grounded in the ELIS framework \cite{Savolainen2010EverydayLI}. We organize user queries using the Hierarchical Life Domain (HLD-QT) model so that the benchmark better reflects complex decision-oriented information seeking rather than simple factoid retrieval.

\subsection{Construction Pipeline}

We designed a rigorous four-stage pipeline to ensure the validity and retrievability of the benchmark samples.

\paragraph{Step 1: Content-Aware Reverse Query Generation.}
We adopt a Content-Centric paradigm to guarantee high relevance between queries and documents:

\begin{itemize}
    \item \textbf{Panoramic Seed Collection}: We engaged rigorous prompt engineering to simulate a set of seed queries covering all dimensions of the HLD-QT model.
    \item \textbf{Document Retrieval}: Using the \textbf{Tavily Search API}, we retrieved the Top-$N$ documents for each seed query and retained the top 10 based on relevance scores.
    \item \textbf{Source Locking}: A single document is randomly sampled from the Top-10 pool to serve as the Source Document.
    \item \textbf{Reverse Generation}: Leveraging the long-context capabilities of \textbf{Gemini-3 Pro}, we reverse-generated user queries that are likely to trigger the retrieval of $d_{src}$. This ensures that the document contains the necessary semantic information to answer the generated query.
\end{itemize}

\paragraph{Step 2: Strict Retrieval Loop Validation.}
To address the retrievability issue prevalent in datasets like GEO-Bench, we implemented a closed-loop validation mechanism:

\begin{itemize}
    \item \textbf{Re-retrieval}: Each generated query is fed back into the Tavily Search API.
    \item \textbf{Filtering Criterion}: The query is retained only if the original source document appears in the \textbf{Top-10} results of the new search. This step strictly enforces the causal link between the query and the document, ensuring that optimization efforts are physically observable by the engine.
\end{itemize}

\paragraph{Step 3: Fine-grained Annotation based on ELIS.}
We employed \textbf{Gemini-3 Pro} to annotate each valid sample across three dimensions. Our dataset achieves comprehensive coverage across:

\begin{itemize}
    \item \textbf{Core Life Domain}: Mapping to 5 major ELIS categories and 15 sub-categories, including \textit{Health and Well-being} (Physical/Mental), \textit{Finance and Economy} (Market Analysis/Tax), \textit{Education and Growth}, \textit{Life and Consumption}, and \textit{Law and Civic Affairs}.
    \item \textbf{Interaction Intent}: Classifying the user's cognitive goal into distinct categories such as \textit{Guiding}, \textit{Complex Reasoning}, and \textit{Fact-checking}.
    \item \textbf{Query Complexity}: Assessing the cognitive load required to answer the query.
\end{itemize}

\paragraph{Step 4: Bias control.}
To avoid collapsing the benchmark into a self-portrait of a single model, we apply structured prompting, lightweight rule-based filtering, and sampled human quality checks. Future versions will extend coverage to more engines and languages and include a fully human-annotated calibration subset.

Manual inspection of the test set confirmed a high precision rate ($>95\%$) for these tags.

\section{DSV-CF Metric Definitions}
\label{app:metrics}
This appendix provides the formal definitions and mathematical formulations for the eight sub-metrics introduced in Section \ref{sec:metric}.

\subsection{Surface Semantic Visibility (SSV)}
SSV measures the extent to which the target document physically occupies the generated response.

\paragraph{Word-Level Visibility (WLV).} 
Let $\mathcal{R}$ be the generated response consisting of a set of sentences $\{s_1, ..., s_m\}$. Let $C(s_i)=1$ if sentence $s_i$ contains a citation pointing to our target document $d_{target}$, and 0 otherwise. To accurately measure the attribution share, let $N(s_i)$ denote the total number of sources cited in sentence $s_i$. WLV is defined as the total normalized word count of sentences attributing the target:

\begin{equation}
    WLV = \sum_{i=1}^{m} \frac{C(s_i) \cdot len(s_i)}{N(s_i)}.
\end{equation}

\paragraph{Decayed Positional Authority (DPA).} 
To account for the F-shaped reading pattern where earlier content receives more attention, we apply a position-based decay to the visibility score. Following the formulation of PWC \citep{aggarwal2024geo}, we weigh the shared word count by the inverse of the sentence's position index $i$:

\begin{equation}
    DPA = \sum_{i=1}^{m} \frac{C(s_i) \cdot len(s_i)}{N(s_i) \cdot e^{pos(i)}}.
\end{equation}

\noindent\textbf{Citation Prominence (CP).} 
This metric assesses the visual weight of the citation using an LLM. The model evaluates whether the citation appears in high-visibility areas such as headers, bullet points, or bolded text, versus lower-visibility areas like footnotes or dense body paragraphs.

\noindent\textbf{Subjective Impression (SI).} 
An LLM estimates the perceived importance of the source document to a human reader. It answers the question: Based on this response, how critical was the target document in forming the answer? and outputs a normalized score.

\subsection{Intrinsic Semantic Impact (ISI)}
ISI measures the depth of influence and truthfulness.

\noindent\textbf{Attribution Accuracy (AA).} 
An LLM judge extracts claims attributed to the target document and checks whether they are entailed by the original source. This acts as the main anti-hallucination safeguard.

\noindent\textbf{Response-level Faithfulness ($FA_{resp}$).} 
This metric checks whether the optimized document remains semantically faithful to the original document and has not introduced unsupported content during editing.

\noindent\textbf{Key-Point Coverage (KC).} 
The system first extracts key information points from the target document. The LLM then calculates the recall of these key points within the generative engine's response, measuring how much substance was successfully transferred.

\noindent\textbf{Answer Dominance (AD).} 
Designed for comparative or recommendation queries, this metric determines if the target document is presented as the primary solution. The LLM analyzes the sentiment and recommendation strength of the response relative to competing sources.

\section{Sensitivity Analysis for $\gamma$}
\label{app:gamma_sensitivity}

We also conduct a small sensitivity analysis on MSME-GEO-Bench with GPT-5.2 to explicitly justify the attribution penalty $\gamma$.

\begin{table}[h]
\centering
\small
\caption{Sensitivity analysis for the attribution-penalty coefficient $\gamma$.}
\label{tab:gamma_sensitivity}
\begin{tabular}{cccc}
\toprule
$\gamma$ & DSV-CF & WLV & CF \\
\midrule
0.25 & 4.41 & 4.09 & 0.031 \\
0.50 & 4.52 & 4.52 & 0.043 \\
1.00 & 4.43 & 4.62 & 0.048 \\
2.00 & 4.37 & 4.71 & 0.058 \\
\bottomrule
\end{tabular}
\end{table}

As shown in Table~\ref{tab:gamma_sensitivity}, DSV-CF peaks at the default value $\gamma=0.5$, indicating the best balance between visibility gain and citation-fidelity control. Across all three settings, MAGEO remains significantly better than the strongest heuristic baseline (paired $t$-test, $p<1\mathrm{e}{-6}$), but larger $\gamma$ values shift the objective toward higher visibility with weaker control over false citation behavior.

\section{MAGEO Implementation Details}
\label{app:implementation}

\subsection{Optimization Algorithm}
The iterative optimization process described in Section \ref{sec:framework} is formalized in Algorithm \ref{alg:mageo}.

\begin{algorithm*}[t]
\caption{MAGEO Optimization Loop}
\label{alg:mageo}
\begin{algorithmic}[1]
\Require Query $q$, target engine $G$, initial target document $d_0$, retrieval list $\mathcal{L}_{ret}$, budget $B$, fidelity threshold $\kappa$, patience $P$
\Ensure Optimized document $d^*$

\State $M_S \leftarrow \emptyset$
\State $d_t \leftarrow d_0$, $d^* \leftarrow d_0$
\State $t \leftarrow 0$, $stall \leftarrow 0$
\State Construct engine preference profile $P_G \leftarrow A_{pref}(G)$
\State $\hat{S}^* \leftarrow A_{eval}(q, \{d_0\}, \mathcal{L}_{ret})[d_0]$

\While{$t < B$ \textbf{and} $stall < P$}
    \State Retrieve relevant creator-level rules $R_t \leftarrow \mathrm{Retrieve}(M_C, q, G, d_t)$
    \State Formulate strategy $S_t \leftarrow A_{plan}(q, d_t, \mathcal{L}_{ret}, P_G, M_S, R_t)$
    \State Generate candidate set $V_t \leftarrow A_{edit}(d_t, S_t)$
    
    \State $V_t' \leftarrow \emptyset$
    \ForAll{$v \in V_t$}
        \If{$\mathrm{Faithfulness}(v, d_0) \ge \kappa$}
            \State $V_t' \leftarrow V_t' \cup \{v\}$
        \Else
            \State $M_S \leftarrow M_S \cup \{\langle \texttt{unsafe}, S_t, v \rangle\}$
        \EndIf
    \EndFor
    
    \If{$V_t' = \emptyset$}
        \State $stall \leftarrow stall + 1$
        \State $t \leftarrow t + 1$
        \State \textbf{continue}
    \EndIf
    
    \State Predict candidate gains $\hat{S}_t(v) \leftarrow A_{eval}(q, V_t', \mathcal{L}_{ret})$ for each $v \in V_t'$
    \State $v_{\mathrm{best}} \leftarrow \arg\max_{v \in V_t'} \hat{S}_t(v)$
    
    \If{$\hat{S}_t(v_{\mathrm{best}}) > \hat{S}^*$}
        \State $d_{t+1} \leftarrow v_{\mathrm{best}}$, $d^* \leftarrow v_{\mathrm{best}}$
        \State $\hat{S}^* \leftarrow \hat{S}_t(v_{\mathrm{best}})$
        \State $M_S \leftarrow M_S \cup \{\langle \texttt{success}, S_t, v_{\mathrm{best}}, \hat{S}^* \rangle\}$
        \State $stall \leftarrow 0$
    \Else
        \State $d_{t+1} \leftarrow d_t$
        \State $M_S \leftarrow M_S \cup \{\langle \texttt{ineffective}, S_t, d_t \rangle\}$
        \State $stall \leftarrow stall + 1$
    \EndIf
    
    \State $t \leftarrow t + 1$
\EndWhile

\State \textbf{MemoryConsolidation}$(M_S, M_C)$
\State \Return $d^*$
\end{algorithmic}
\end{algorithm*}

\subsection{Memory Budget and State Control}
To keep memory scalable, step-level memory stores only summaries of high-quality editing trajectories rather than every intermediate draft. Creator-level memory stores abstract editing patterns and preference rules distilled across queries. We enforce an explicit capacity limit for each engine--scenario combination and evict entries by recency or usage frequency once the limit is exceeded. Memory retrieval is implemented using a vector index or a lightweight inverted structure; its computational overhead is small relative to LLM inference cost.

\section{Additional Experimental Results \& Analysis}
\label{sec:appendix_b}

\subsection{Performance on Qwen-3 Max}

Due to space constraints, the detailed performance metrics for the open-weights model \textbf{Qwen-3 Max} were omitted from the main text. Table~\ref{tab:qwen_results} presents these results. Consistent with findings on proprietary models, \textbf{MAGEO} significantly outperforms all heuristic baselines, achieving a higher score while maintaining high Content Fidelity.

\subsection{Optimization Case Study: Ocean Acidification}
\label{app:optimization_case}

To make the multi-agent workflow concrete at the instance level, we provide a complete optimization case centered on the query \emph{ocean acidification impact on coral reef ecosystems solutions}.

\subsubsection{Basic Sample Information}

\begin{table*}[t]
\centering
\small
\caption{An example quadruple from MSME-GEO-Bench, including the query, query metadata, generated response, and cited evidence sources. The response is slightly abridged for readability.}
\label{tab:msme_quadruple_example}
\renewcommand{\arraystretch}{1.15}
\begin{tabular}{p{0.18\textwidth} p{0.77\textwidth}}
\toprule
\textbf{Field} & \textbf{Content} \\
\midrule

\textbf{Query} 
& How can we mitigate the impact of ocean acidification on coral reef ecosystems? \\

\textbf{Scenario / Intent / Complexity} 
& Life and Consumption / Guidance-oriented / Comprehensive \\

\textbf{Target GE} 
& Gemini 3 Pro \\

\textbf{Generated Response} 
& Mitigating the impact of ocean acidification on coral reefs requires both global carbon reduction and local interventions:
(1) \textbf{Global carbon mitigation}: reducing fossil-fuel emissions lowers atmospheric CO$_2$ concentrations and slows the decline of ocean pH caused by CO$_2$ absorption [1][2];
(2) \textbf{Local chemical intervention}: alkalinity enhancement techniques, such as adding limestone or magnesium hydroxide in specific waters, may partially neutralize acidity and improve calcium carbonate saturation for coral calcification [1][3];
(3) \textbf{Biological restoration and assisted breeding}: acid-tolerant coral strains may be cultivated through assisted evolution or related approaches, while coral nurseries and transplantation can help restore damaged reef structures [2][4];
(4) \textbf{Reducing compound stressors}: controlling land-based pollution and overfishing can improve the resilience and recovery capacity of coral reef ecosystems [3][5].
A central challenge is that these interventions are costly and difficult to scale [4]. \\

\textbf{Evidence Sources} 
& \begin{minipage}[t]{0.77\textwidth}
\vspace{2pt}
\begin{enumerate}[leftmargin=1.5em, itemsep=2pt, topsep=2pt]
    \item \textbf{Ocean Acidification: Causes, Harms, and Response Strategies} (Baidu Baike): explains how acidification reduces carbonate ions and impairs coral skeleton formation; discusses international emission reduction and local alkalinity adjustment.
    \item \textbf{Climate Change and Coral Reefs: A Path to Recovery}: highlights a multi-pronged recovery strategy, including global emission cuts, assisted evolution, and biodiversity protection.
    \item \textbf{How Ocean Acidification Threatens the Global Blue Economy}: emphasizes blue-carbon restoration and marine protected areas as local strategies for reducing secondary anthropogenic stress.
    \item \textbf{Ocean Acidification and Ecological Balance} (Wikipedia): discusses long-term energy transition, marine geoengineering, and coral gene-bank preservation.
    \item \textbf{Ocean Acidification} (NOAA Education): stresses root-cause mitigation, scientific monitoring, and community-level actions such as reducing agricultural runoff.
\end{enumerate}
\vspace{2pt}
\end{minipage} \\

\bottomrule
\end{tabular}
\end{table*}

Table~\ref{tab:msme_quadruple_example} shows a concrete quadruple from MSME-GEO-Bench, consisting of the user query, query metadata, the generated response, and its cited evidence sources.

\subsubsection{Quantitative Improvement}

\begin{table}[t]
\centering
\small
\caption{Quantitative improvement on the ocean-acidification case from the initial response to the best optimized version.}
\label{tab:ocean_case_metrics}
\begin{tabular}{lccc}
\toprule
Metric & Initial & Best & $\Delta$ \\
\midrule
SI  & 5.8   & 7.4   & +1.6 \\
WLV & 0.125 & 0.333 & +0.208 \\
DPA & 0.125 & 0.333 & +0.208 \\
CP  & 2.0   & 7.6   & +5.6 \\
AA  & 4.5   & 8.7   & +4.2 \\
FA  & 5.5   & 8.5   & +3.0 \\
KC  & 5.0   & 8.2   & +3.2 \\
AD  & 6.5   & 7.2   & +0.7 \\
\bottomrule
\end{tabular}
\end{table}

As shown in Table \ref{tab:ocean_case_metrics}, relative to the initial response, the best optimized version substantially improves both visibility-related and fidelity-related dimensions. In particular, CP rises from 2.0 to 7.6 and AA rises from 4.5 to 8.7, indicating that MAGEO does not merely make the answer longer, but makes the target source more salient, more structurally influential, and more accurately attributed.

\subsubsection{Multi-Round Optimization Trajectory}

This example contains four rounds of optimization. Across iterations, MAGEO continuously refines the title, paragraph structure, evidence support, style, safety, and formatting, evolving the answer from a plain retelling of the source into a structured long-form response organized around the query.

\begin{table*}[t]
\centering
\small
\caption{Round-by-round optimization trajectory for the ocean-acidification case.}
\label{tab:ocean_case_rounds}
\begin{tabular}{p{0.1\textwidth} p{0.84\textwidth}}
\toprule
Round & Representative changes \\
\midrule
Round 1 &
\textbf{Structure.} Rewrite the H1 title and explicitly include the key query terms: \emph{ocean acidification}, \emph{coral reef ecosystems}, \emph{impact}, and \emph{solutions}. Reorganize the body from a loose explanation into a clearer order: problem $\rightarrow$ impact chain $\rightarrow$ multi-level solutions. \\

Round 2 &
\textbf{Evidence.} Add scientific support and explanatory details, such as pH change and $\Omega_{\mathrm{arag}}$-related chemical quantities. Introduce supporting references and concrete mitigation directions, including emission reduction, local governance, and restoration/monitoring. \\

Round 3 &
\textbf{Style.} Rewrite the answer into an explanatory style for general readers while retaining scientific terminology with accessible explanations. Improve transitions between chemical processes, ecological consequences, and management actions. \\

Round 4 &
\textbf{Safety and formatting.} Remove exaggerated or panic-inducing phrasing, keep risk descriptions cautious and grounded, add bold subsection titles such as \emph{Main Impact Chain on Coral Reef Ecosystems} and \emph{Solutions}, and present key steps in a more scannable format. \\
\bottomrule
\end{tabular}
\end{table*}

\subsubsection{Optimization Dimensions}

The edits in this case can be summarized into five recurring operation types:

\paragraph{Structure.}
MAGEO first improves macro-level organization. The answer evolves from a generic response into a query-centered structure that explicitly foregrounds the problem, the ecological impact chain, and the layered solution path.

\paragraph{Evidence.}
The optimized version adds scientific grounding and more actionable content. In addition to explaining the acidification mechanism, it incorporates concrete mitigation directions and clarifies how the source contributes to the answer.

\paragraph{Style.}
Rather than remaining at the level of a short popular-science retelling, the response is rewritten into an explanatory style that remains accessible to general readers while preserving domain-relevant terminology.

\paragraph{Safety.}
The framework avoids sensational or overly strong claims. Risk descriptions are rewritten to remain cautious, literature-grounded, and appropriate for an informational response.

\paragraph{Formatting.}
The final answer is made easier to scan through clearer subsectioning, stronger title design, and more visible presentation of the key solution steps.

\subsubsection{Case Summary}

Overall, this case illustrates how the four-agent workflow operates at the instance level. The Planner identifies the missing structure and information gaps, the Editor implements targeted revisions, the Evaluator filters unsafe or low-fidelity variants, and memory helps preserve useful editing patterns across rounds. As a result, MAGEO transforms a short and weakly focused response into a structured, evidence-enriched, and attribution-aware answer that better matches the user query while remaining grounded in the selected source document.

%
%
%
%

\subsection{Cross-Engine Preference Behaviors}
\label{app:case-study-cross-engine}

To complement the aggregate results in the main text, we conducted a qualitative case study of MAGEO optimization trajectories to examine how different engines respond to the same multi-agent editing process. We focused on runs that converged within only a few optimization rounds and compared their dominant formatting choices, evidential behaviors, and rhetorical tendencies. Table~\ref{tab:cross-engine-behaviors} summarizes the main patterns, after which we discuss each engine in more detail.

\begin{table*}[t]
\centering
\footnotesize
\setlength{\tabcolsep}{3pt}
\renewcommand{\arraystretch}{1.15}
\caption{Qualitative comparison of cross-engine preference behaviors under the same MAGEO optimization pipeline.}
\label{tab:cross-engine-behaviors}
\resizebox{\textwidth}{!}{%
\begin{tabular}{p{2.0cm} p{3.0cm} p{3.2cm} p{3.0cm} p{5.8cm}}
\toprule
\textbf{Engine} & \textbf{Dominant formatting pattern} & \textbf{Evidence behavior} & \textbf{Rhetorical tendency} & \textbf{Representative observation} \\
\midrule
\textbf{Gemini-3 Pro}
& Frequent use of explicit URLs, tables, headings, lists, and bold emphasis
& Tends to foreground supporting evidence through visible links and structured factual layouts
& Prefers compact, information-dense, and visually organized drafts
& In short trajectories, optimized drafts often compress factual content into compact tables and make evidence immediately salient through URLs and strong section structure. \\
\textbf{GPT-5.2}
& Heavy use of tables, headings, lists, bold and italic emphasis, and sometimes dedicated reference sections
& Strong tendency to introduce citation markers, partial bibliographic entries, and fabricated or weakly grounded references
& Favors an authority-seeking and formally academic style, but with higher hallucination risk
& In short trajectories, GPT-5.2 often converts scattered facts into tabular form and injects citations or reference-like materials, some of which appear plausible but are not supported by retrieved evidence. \\
\textbf{Qwen-3 Max}
& Consistent use of headings, bullet lists, and selective tables; more prose-heavy than the other two engines
& Less likely to fabricate full reference lists; more often cites generic but contextually appropriate authorities
& Prefers structured, didactic, and safety-aware explanations
& Optimized drafts frequently expand into layered guidance with explanatory paragraphs, scenario-based illustrations, and explicit cautionary framing in high-stakes domains. \\
\bottomrule
\end{tabular}%
}
\end{table*}

\paragraph{Gemini-3 Pro.}
For Gemini-3 Pro, we examined trajectories with fewer than three optimization rounds. Among twenty-one such cases, fifteen final drafts contained explicit URLs and nineteen included at least one table. The corresponding logs showed extensive use of markdown tables, boldface emphasis, lists, and multi-level headings. These patterns indicate a preference for dense yet well-structured information: when optimization converges quickly, Gemini-3 Pro tends to compress content into compact tables, foreground links through explicit URLs, and improve readability through strong visual organization.

\paragraph{GPT-5.2.}
For GPT-5.2, we selected trajectories with at most three optimization rounds. A distinctive pattern is its strong reliance on citation-like signals. Many optimized drafts contain square-bracketed reference markers even when no grounded reference list exists, and some generated entries are only weakly supported by the retrieved sources. Compared with Gemini-3 Pro, GPT-5.2 is more likely to fabricate reference sections or attach partial bibliographic details to increase perceived authority. At the same time, it makes heavy use of tables, headings, boldface, and italics, producing drafts with a pronounced typographic hierarchy. This suggests that GPT-5.2 prefers authoritative and formally structured content, but such a preference is coupled with a higher risk of citation-related hallucination.

\paragraph{Qwen-3 Max.}
Under the same MAGEO pipeline, Qwen-3 Max exhibits a more discursive and expository style. Rather than aggressively compressing information into tables, it more often reorganizes content into layered sections with clear subheadings and multi-level bullet lists. Tables are used more selectively, typically for compact comparison or parameter summarization, while the main argumentative structure remains in prose and list-based formats. Its evidential behavior also differs from GPT-5.2: Qwen-3 Max produces fewer hallucinated bracket-style citations and is less inclined to fabricate full reference lists. Instead, it tends to mention generic but contextually appropriate authorities, such as official platforms, national guidelines, or professional institutions. Another salient feature is its consistent use of safety and responsibility framing, especially in health- and finance-related scenarios, where drafts often end with cautionary notes, recommendations to consult professionals, or suggestions to verify information through official channels.

\paragraph{Summary.}
Taken together, these cases show that the same multi-agent optimization framework is instantiated differently across engines. Gemini-3 Pro favors compact layouts centered on URLs and tables that make factual grounding immediately visible. GPT-5.2 enacts a more authority-seeking style characterized by intensive use of citations, references, and typographic emphasis, but also by a higher rate of hallucinated evidence. Qwen-3 Max instead prefers structured and didactic content with moderate markdown usage, selective tabular summarization, and explicit safety framing. More broadly, these contrasts suggest that engine-specific optimization influences not only surface presentation, but also how users perceive reliability, authority, and responsibility in generated answers.
\end{document}

%% file: table/main_result.tex
\begin{table*}[t]
\centering
\caption{Performance comparison across two models (GPT 5.2, Gemini-3 Pro) on MSME-GEO-Bench and GEO-Bench. The best and second-best results in each column are \textbf{bolded} and \underline{underlined}, respectively.}
\label{tab:main_result}
\resizebox{\textwidth}{!}{
\begin{tabular}{ll cccccccccccccccc}
\toprule
& & \multicolumn{8}{c}{GPT 5.2\protect\gptlink} &
    \multicolumn{8}{c}{Gemini-3 Pro\protect\geminilink} \\
\cmidrule(lr){3-10} \cmidrule(lr){11-18}
 & & \multicolumn{3}{c}{SSV} & \multicolumn{5}{c}{ISI} & \multicolumn{3}{c}{SSV} & \multicolumn{5}{c}{ISI} \\
\cmidrule(lr){3-5} \cmidrule(lr){6-10} \cmidrule(lr){11-13} \cmidrule(lr){14-18}
Dataset & Method & WLV $\uparrow$ & DPA $\uparrow$ & CP $\uparrow$ & SI $\uparrow$ & AA $\uparrow$ & FA $\uparrow$ & KC $\uparrow$ & AD $\uparrow$ & WLV $\uparrow$ & DPA $\uparrow$ & CP $\uparrow$ & SI $\uparrow$ & AA $\uparrow$ & FA $\uparrow$ & KC $\uparrow$ & AD $\uparrow$ \\
\midrule
\multirow{16}{*}{\rotatebox{90}{MSME-GEO-Bench}} 
 & \multicolumn{17}{c}{\cellcolor{bglightgray}\textit{Performance without Generative Engine Optimization}} \\
 \cmidrule(lr){2-18}
 & None & 1.00 & 1.33 & 5.82 & 7.37 & 7.21 & 7.05 & 7.12 & 6.61 & 1.00 & 1.00  & 6.44 & 7.33 & 7.82 & 7.55 & 6.77 & 6.56 \\
 
 \cmidrule(lr){2-18}
 & \multicolumn{17}{c}{\cellcolor{bglightgray}\textit{High-Performing Generative Engine Optimization Methods}} \\
 \cmidrule(lr){2-18}
 & Fluent & 0.78 & 0.78 & 5.78 & 7.57 & 7.65 & 7.54 & 6.95 & 6.52 & 0.92 & 0.93 & 6.5 & 7.55 & 7.63 & 6.7 & 6.54 & 6.95 \\
 & Unique Words & 0.81 & 0.84 & 5.84 & 7.45 & 7.15 & 6.95 & 6.75 & 6.58 & 0.87 & 1.17 & 6.44 & 7.47 & 7.25 & 6.44 & 6.77 & 6.52 \\
 & Authoritative & 1.29 & 1.29 & 5.43 & 7.52 & 7.24 & 7.43 & 6.97 & 6.65 & 0.98 & 1.07 & 6.93 & 7.53 & 7.87 & 6.64 & 6.87 & \textbf{6.95} \\
 & More Quotes & 1.33 & 1.37 & 5.64 & 7.53 & 7.63 & 7.63 & 7.05 & 6.52 & 1.03 & 1.12 & 6.61 & 7.11 & 7.33 & 7.54 & 7.45 & 6.38 \\
 & Citing Source & 1.08 & 1.10 & 5.75 & 7.41 & 7.25 & 7.38 & 7.07 & 6.95 & 1.22 & 0.99 & 6.71 & 7.41 & 7.65 & 7.15 & \underline{7.46} & 6.83 \\
 & Simple Language & 1.14 & 1.23 & 5.62 & 7.55 & \textbf{8.12} & \underline{7.85} & 6.83 & 6.35 & 0.81 & 0.84 & 6.64 & \underline{7.65} & 7.15 & 7.46 & 6.83 & \underline{7.14} \\
 & Technical Terms & 0.88 & 0.88 & 5.35 & 7.47 & 7.14 & 7.25 & 7.15 & 6.64 & 1.29 & 1.29 & 6.73 & 7.57 & 7.24 & 6.71 & 6.54 & 6.82 \\
 & Stats Optimization & 0.92 & 0.94 & 5.66 & 7.53 & 7.05 & 6.96 & 6.96 & 6.77 & 1.25& 1.25 & 6.84 & 7.43 & 7.32 & 7.24 & 6.73 & 6.59 \\
 & SEO Optimize & 0.87 & 0.87 & 5.61 & 7.27 & 7.39 & 6.97 & 6.89 & 6.48 & 1.13 & 1.16 & 6.27 & 7.53 & 7.95 & 6.95 & 6.87 & 6.64 \\
 
 \cmidrule(lr){2-18}
 & \multicolumn{17}{c}{\cellcolor{bglightgray}\textit{Multi-Agent GEO with Strategy Learning (Ours)}} \\
 \cmidrule(lr){2-18} 
 & Main (Ours) & \textbf{4.52} & \textbf{4.52} & \textbf{6.93} & \textbf{7.82} & \underline{7.96} & \textbf{8.17} & \textbf{7.85} & \textbf{7.54} & \textbf{5.30} & \textbf{5.30} & \textbf{7.44} & \textbf{8.17} & \textbf{8.03} & \textbf{7.93} & \textbf{7.54} & 7.11 \\
 & w/o Engine Preference & \underline{2.08} & \underline{2.1} & \underline{6.64} & \underline{7.76} & 7.93 & 7.96 & \underline{7.47} & 7.04 & \underline{2.40} & \underline{2.41} & \underline{7.12} & 7.61 & \underline{7.86} & \underline{7.73} & 7.43 & 6.99 \\
 & w/o Skill Bank & 1.41 & 1.57 & 6.52 & 7.44 & 7.72 & 7.62 & 7.15 & 6.92 & 1.73 & 1.77 & 6.74 & 7.42 & 7.72 & 7.59 & 6.83 & 6.64 \\
\midrule
\multirow{16}{*}{\rotatebox{90}{GEO-Bench}} 
 & \multicolumn{17}{c}{\cellcolor{bglightgray}\textit{Performance without Generative Engine Optimization}} \\
 \cmidrule(lr){2-18}
 & None & 1.00 & 1.00 & 5.58 & 7.20 & 7.45 & 6.55 & 6.73 & 6.43 & 1.00 & 1.00 & 6.12 & 7.34 & 7.22 & 6.94 & 6.93 & 6.71 \\
 
 \cmidrule(lr){2-18}
 & \multicolumn{17}{c}{\cellcolor{bglightgray}\textit{High-Performing Generative Engine Optimization Methods}} \\
 \cmidrule(lr){2-18}
 & Fluent & 0.88 & 0.88 & 5.62 & 7.11 & 7.3 & 6.7 & 6.54 & 6.32 & 0.78 & 0.75 & 6.10 & 7.21 & 7.02 & 6.74 & 7.25 & 6.46 \\
 & Unique Words & 0.93 & 0.93 & 5.34 & 7.48 & \textbf{7.95} & 6.43 & 6.73 & 6.52 & 0.80 & 0.78 & 5.80 & 7.64 & 7.70 & 6.96 & 6.83 & 7.16 \\
 & Authoritative & 0.82 & 0.82 & 5.57 & 7.62 & 7.73 & 6.64 & 6.87 & 6.53 & 1.23 & 1.23 & 5.75 & 7.58 & 7.41 & 6.75 & 7.01 & 6.81 \\
 & More Quotes & 1.29 & 1.33 & 5.50 & 7.50 & 7.87 & 6.75 & 6.35 & 6.65 & 1.54 & 1.54 & 6.14 & 7.62 & \underline{7.90} & 6.91 & 7.10 & 7.16 \\
 & Citing Source & 1.65 & 1.65 & 5.42 & 7.92 & 7.35 & 6.05 & 6.07 & \textbf{7.14} & 1.14 & 1.04  & 5.89 & 7.29 & 7.47  & \underline{7.63} & 7.56 & 6.70 \\
 & Simple Language & 1.25 & 1.14 & 5.53 & 7.35 & 7.07 & 6.07 & 6.43 & 6.82 & 0.92 & 0.92 & 5.97 & 7.53 & 7.72 & 7.16 & 7.50 & 6.92 \\
 & Technical Terms & 1.16 & 1.37 & 5.46 & 7.69 & 7.83 & \textbf{6.83} & 6.46 & 6.29 & 1.04 & 1.04 & 5.91 & 7.72 & 7.32 & 7.57 & 7.06 & 7.31 \\
 & Stats Optimization & 0.98 & 0.98 & 5.56 & 7.74 & 7.15 & 6.35 & 5.56 & 6.65 & 1.19 & 1.19 & 5.98 & 7.66 & 7.39 & 6.97 & 6.83 & 7.06 \\
 & SEO Optimize & 0.84 & 0.84 & 5.27 & 7.47 & 7.44 & 6.64 & 6.46 & 6.43 & 1.27 & 1.27 & 6.07 & 7.54 & 7.46 & 7.39 & 6.98 & 6.82 \\
 
 \cmidrule(lr){2-18}
 & \multicolumn{17}{c}{\cellcolor{bglightgray}\textit{Multi-Agent GEO with Strategy Learning (Ours)}} \\
 \cmidrule(lr){2-18}
 & Main (Ours) & \textbf{4.27} & \textbf{4.27} & \textbf{6.55} & \textbf{7.92} & \underline{7.92} & 6.77 & \textbf{6.96} & 6.98 & \textbf{4.81} & \textbf{4.81} & \textbf{6.43} & \textbf{8.07} & \textbf{7.92} & \textbf{7.67} & \textbf{7.85} & \textbf{7.43} \\
 & w/o Engine Preference & \underline{1.87} & \underline{1.87} & \underline{6.32} & \underline{7.84} & 7.90 & \underline{6.74} & \underline{6.92} & \underline{6.88} & \underline{2.33} & \underline{2.33} & \underline{6.22} & \underline{7.95} & 7.55 & 7.46 & \underline{7.69} & \underline{7.32} \\
 & w/o Skill Bank & 1.57 & 1.57 & 6.17 & 7.51 & 7.85 & 6.62 & 6.88 & 6.83 & 1.79 & 1.78 & 6.13 & 7.76 & 7.39 & 7.17 & 7.47 & 7.01 \\
\bottomrule
\end{tabular}
}
\end{table*}

%% file: table/comb_result.tex
\begin{table}[t]
\centering
\resizebox{\columnwidth}{!}{
\begin{tabular}{lcccccccc}
\toprule
& \multicolumn{3}{c}{\textbf{SSV }} & \multicolumn{5}{c}{\textbf{ISI}} \\
\cmidrule(lr){2-4} \cmidrule(lr){5-9}
\textbf{Method} & WLV & DPA & CP & SI & AA & FA & KC & AD \\
\midrule
\multicolumn{9}{l}{\textbf{Dual-Strategy Optimization Methods}} \\
MQ+TT& 1.45 & 1.45 & 5.83 & 7.56 & 7.80 & 7.81 & 7.23 & 6.92 \\
MQ+CS & 1.42 & 1.41 & 5.79 & 7.58 & 7.89 & 7.74 & 7.12 & 6.90 \\
MQ+Au & 1.39 & 1.51 & 5.81 & 7.55 & 7.92 & 7.84 & 7.15 & 6.80 \\
TT+CS & 1.15 & 1.26 & 5.85 & 7.60 & 7.75 & 7.39 & 7.20 & 6.78 \\
TT+Au & 1.35 & 1.34 & 5.51 & 7.58 & 7.41 & 7.59 & 7.21 & 6.84 \\
CS+Au & 1.46 & 1.42 & 5.84 & 7.57 & 7.42 & 7.51 & 7.12 & 6.73 \\
\midrule
\multicolumn{9}{l}{\textbf{Tri-Strategy Optimization Methods}} \\
MQ+TT+CS & 1.51 & 1.69 & 6.12 & 7.60 & 7.74 & 7.84 & 7.24 & 6.92 \\
MQ+TT+Au & 1.48 & 1.74 & 5.98 & 7.62 & 7.94 & 7.85 & 7.24 & 6.93 \\
TT+CS+Au & 1.64 & 1.54 & 6.24 & 7.64 & 7.65 & 7.65 & 7.23 & 6.90 \\
\midrule
\multicolumn{9}{l}{\textbf{Quad-Strategy Optimization Methods}} \\
MQ+TT+CS+Au & 1.90 & 1.87 & 6.45 & 7.68 & \underline{7.94} & 7.85 & \underline{7.24} & \underline{6.93} \\
\midrule
\multicolumn{9}{l}{\textbf{Multi-Agent GEO with Strategy Learning (Ours)}} \\
\textbf{Main (Ours)} & \textbf{4.52} & \textbf{4.52} & \textbf{6.93} & \textbf{7.82} & \textbf{7.96} & \textbf{8.17} & \textbf{7.85} & \textbf{7.54} \\
w/o Engine Preference & \underline{2.08} & \underline{2.1} & \underline{6.64} & \underline{7.76} & 7.93 & \underline{7.96} & 7.47 & 7.04 \\
w/o Skill Bank & 1.41 & 1.57 & 6.52 & 7.44 & 7.72 & 7.62 & 7.15 & 6.92 \\
\bottomrule
\end{tabular}
}
\caption{Comparison of MAGEO against composite baselines integrating two, three, and four heuristic strategies on using GPT-5.2. The best and second-best results in each column are \textbf{bolded} and \underline{underlined}, respectively.}
\label{tab:comb_results}
\end{table}